# The Hidden Vulnerability of Distributed Learning in Byzantium

El Mahdi El Mhamdi [1]   Rachid Guerraoui [1]   Sébastien Rouault [1]

## Abstract

While machine learning is going through an era of celebrated success, concerns have been raised about the vulnerability of its backbone: stochastic gradient descent (SGD). Recent approaches have been proposed to ensure the robustness of distributed SGD against adversarial (Byzantine) workers sending *poisoned* gradients during the training phase. Some of these approaches have been proven *Byzantine–resilient*: they ensure the *convergence* of SGD despite the presence of a minority of adversarial workers. We show in this paper that *convergence is not enough*. In high dimension $d \gg 1$, an adversary can build on the loss function's non–convexity to make SGD converge to *ineffective* models. More precisely, we bring to light that existing Byzantine–resilient schemes leave a *margin of poisoning* of $\Omega(f(d))$, where $f(d)$ increases at least like $\sqrt{d}$. Based on this *leeway*, we build a simple attack, and experimentally show its strong to utmost effectivity on CIFAR–10 and MNIST. We introduce *Bulyan*, and prove it significantly reduces the attacker's leeway to a narrow $\mathcal{O}(1/\sqrt{d})$ bound. We empirically show that Bulyan does not suffer the fragility of existing aggregation rules and, at a reasonable cost in terms of required batch size, achieves convergence *as if* only non–Byzantine gradients had been used to update the model.

## 1. Introduction

Stochastic Gradient Descent (SGD), is arguably the backbone of the most successful machine learning methods (LeCun et al., 2015; Abadi et al., 2016; Dean et al., 2012; Bottou, 1998). Gradient Descent (GD), the underlying principle of SGD, is so straightforward that, as sometimes said, "Newton could have invented it in his time". In particular, GD relies on a simple observation: given a function $Q$, depending on a parameter $x$, if one keeps updating $x$ in the opposite direction of the gradient of $Q$, with reasonably small, but not too small (Bottou, 1998) steps, $x$ eventually reaches the global minimum if $Q$ is convex or, if $Q$ is not convex[1], reach a region where $Q$ is either flat or in some *local minima*. SGD is the lightweight version of GD, where a sample is drawn at random to estimate the gradient of $Q$.

Beyond image recognition or video labeling for social networks, SGD–based machine learning is venturing into safety–critical applications, like health–care (Holzinger, 2016) and transportation (Bojarski et al., 2016). Meanwhile, a growing body of work, coined *Adversarial Machine Learning* (Biggio & Roli, 2017; Goodfellow et al., 2014; Gilmer et al., 2018; Kumar et al., 2017) is unveiling serious vulnerabilities in some of the most performing algorithms. Essentially, the general effort towards robust ML is conducted against three kinds of attacks: *poisoning* ones (Biggio & Laskov, 2012; Koh & Liang, 2017) where an adversary injects poisoned data during the training phase, *exploratory* attacks, where a curious attacker attempts to infer privacy–sensitive information, and *evasion* attacks, where attackers try to fool an already trained model with adversarial inputs. The three fronts are complementary and each kind of attack poses a challenge on its own.

In the context of poisoning attacks, an emerging line of research looks at robustness through the lenses of (distributed) optimization (Chen et al., 2017; Su, 2017; Blanchard et al., 2017). Interestingly, SGD can be proven to converge despite the presence of a (bounded) number of adversaries. The general recipe is to find a robust estimator for the gradient in the presence of adversaries, and to prefer that estimator over a mere linear combination of the provided gradients (Dean et al., 2012; Zhang et al., 2015), known for not being robust (Rousseeuw, 1985).

To scale, ML implementations are distributed (Abadi et al., 2016; Dean et al., 2012): several workers collaborate to compute a gradient, aggregated by a *parameter server* (Li et al., 2014a;b). In this context, robustness is even more important. A bias in the gradient estimation may not only come from poor sampling or noisy data, but also from *Byzantine* (adversarial) workers (Lamport et al., 1982).

---

[1]EPFL, Lausanne, Switzerland. Correspondence to: (without spaces) <firstname.lastname@epfl.ch>.

[1]In practically interesting cases such as neural networks, $Q$ is far from being convex (Choromanska et al., 2015a;b).

Accordingly, Byzantine–resilient aggregation rules were derived and proved to guarantee the convergence of SGD in this context (Chen et al., 2017; Blanchard et al., 2017). **But is convergence enough** in the non–convex, high dimensional case of neural networks?

In fact, the question of SGD convergence in the case of neural networks is neither new and unprecedented nor old and forgotten by the ML community. Last year, it sparked a hot debate when some, using timely examples, were blaming SGD to be *"brittle"*[2]. Others moderated that view, reminding us how blaming neural networks for their lack of (provable) convergence led the community to *"threw the baby with the bathwater back in the 1990s"*[3]; which arguably slowed research progress in the understanding of neural networks as a learning machine.

We show in the paper that **convergence is not enough**. We look at the question from the robust distributed optimization point of view, where some workers can be Byzantine and we show that, whilst indeed neural networks usually benefit from the existence of many "similarly good" local minima (Choromanska et al., 2015a), making SGD convergence a preferable requirement is clearly not sufficient in a Byzantine distributed setting. More precisely, we show that a single Byzantine worker can make any known Byzantine-resilient aggregation rule for SGD learn *ineffectual* models. It is important to note that we do not contradict the proof of convergence of these rules. We rather take advantage of the high dimensional and highly non–convex *landscape* of the loss function to make these rules *converge*, as they were proven to, but to *ineffectual* models. We provide an analytic understanding of how attackers could benefit from this *curse of dimensionality*, and propose a solution that enhances the Byzantine–resilience of SGD.

To start feeling the vulnerability, consider any Byzantine–resilient gradient aggregation rule based on a distance criteria. When the distance criteria relies solely on norms in the $\ell_p$ categories, with a *small* p (the Euclidean or $\ell_1$ norms), a Byzantine gradient can both remain close to honest gradients and have one of its coordinates *poisoned*, e.g. set to a large value. These poisoned coordinates can take values on the order of $\Omega(\sqrt[p]{d})$, a large order given the dimension $d$ of modern neural networks. And the gradient would still be seen as "legitimate" by the aggregation rule. Inversely, when the distance criteria involves norms closer to the infinite norm, the Byzantine worker can poison every coordinates while being exactly on the value that the aggregation rule will decide on. This way, the Byzantine worker can drive the aggregation to *converge*, but to the worst possible sub-optimum that the $\Omega(\sqrt[p]{d})$ ($\ell_p$ norm) or the $\Omega(d)$

---

[2]Ali Rahimi, Test of Time Award Lecture. NIPS 2017.

[3]Yann LeCun. *"My Take on Ali Rahimi's Test of Time Award Talk at NIPS 2017"*. https://www2.isye.gatech.edu/~tzhao80/Yann_Response.pdf

(infinite norm) margins enable it to get. See section 3.1.

Given any Byzantine–resilient rule $\mathcal{A}$, we propose a generic enhancement recipe we call *Bulyan of* $\mathcal{A}$, or simply *Bulyan*($\mathcal{A}$), that improves Byzantine–resilience in the following sense: if the vectors selected by $\mathcal{A}$ are in a *cone* of angle $\alpha$ around the true gradient, then the vectors selected by *Bulyan*($\mathcal{A}$) are in a *cone* of angle $\alpha' \leq \alpha$. Most importantly, we prove that *Bulyan*($\mathcal{A}$) drastically reduces the *leeway* of Byzantine workers to a narrow $\mathcal{O}(1/\sqrt{d})$ bound.

We also empirically evaluate the trade–offs induced by Bulyan, and compare its convergence speed with those of other aggregation rules. In particular, we show that with typically used values for the batch size, Bulyan is comparable to the speed of averaging (the fastest aggregation rule), which does not stand a single Byzantine worker.

The rest of the paper is organized as follows. Section 2 introduces the model of distributed SGD with Byzantine workers, and recalls Byzantine–resilient gradient aggregation rules. Section 3 introduces our attack and analytically discusses how it affects the aforementioned gradient aggregation rules. Section 4 describes our algorithm, Bulyan, and proves our two main theoretical results. Section 5 highlights the practical impacts of our attack, and the beneficial effects of Bulyan on MNIST and CIFAR–10. Finally, Section 6 concludes by discussing related and future work.

The code used to carry our experiments out (including additional ones asked by reviewers) is available at https://github.com/LPD-EPFL/bulyan.

## 2. Model

### 2.1. Distributed Stochastic Gradient Descent (DSGD)

We follow the usual *parameter server* model used in distributed implementations of machine learning (Dean et al., 2012; Li et al., 2014a;b). The system consists in $n + 1$ processes: 1 *master* and $n$ *workers*. There are $f \leq n$ Byzantine, i.e. adversarial, workers. Their role and capability are described in Section 2.2.

Let $t$ be the current epoch, and $x_t$ be the model parameters, let $Q$ be the cost function we aim to minimize. Each honest worker $i \in \{1, \ldots, n - f\}$ draws i.i.d. (mini–batches of) samples $\xi_i^t$ from the data set to compute an estimate $V_i^t = G(x_t, \xi_i^t)$ of the gradient $\nabla Q(x_t)$.

We assume that $\nabla Q$ is $K$–Lipschitz and that $Q$ is three–times differentiable. We also assume that $G$ has a bounded variance $\sigma$, i.e. $\mathbb{E}_\xi \|G(x, \xi) - \nabla Q(x)\|^2 \leq \sigma^2$, and as assumed in (Bottou, 1998; Blanchard et al., 2017), for $r \in \{1, \ldots, 4\}$, the $r$–th statistical moments of $G$ does not overgrow the $r$–th powers of the model:

$$\exists (A_r, B_r) \in (\mathbb{R}_+)^2, \mathbb{E}_\xi \|G(x, \xi)\|^r \leq A_r + B_r \|x\|^r$$

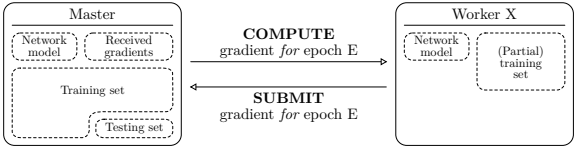

*Figure 1.* A view on the master and a worker X. The most important pieces of information each party hold, along with the communication protocol (COMPUTE and SUBMIT), are represented.

## 2.2. Adversary

The *adversary* in this model is an entity which controls $f$ of the $n$ workers. These adversarial workers, and the gradients they send, are called *Byzantine*. The goal of the adversary is to prevent the distributed SGD process from converging to a *satisfying state*. Ideally, a satisfying state with Byzantine workers should achieve an accuracy that is *comparable to* the one achieved without any Byzantine worker, all other things being equal; e.g. see Figure 4.

The adversary is omniscient, in the sense that it has a perfect knowledge of the *system state* at any time. The system state is constituted exhaustively by:

- the *full* state of the master (data and code)
- the *full* state of every worker (data and code)
- any data exchanged over any communication channel

Hence the adversary can leverage its knowledge of the master's state and the submitted gradients to build powerful attacks, as shown in sections 3.2 and 3.3. However, the adversary is not omnipotent. It cannot *directly* modify the state of the system, impersonate other workers, or delay communications. The adversary is only allowed to submit gradients via its $f$ workers.

## 2.3. Gradient aggregation rules (GAR)

In this section, we describe the three studied *gradient aggregation rule*. A GAR transforms the workers' submitted gradients into a single, aggregated one. This aggregated gradient is then used to update the model parameters.

A very commonly used aggregation rule is *averaging* (Abadi et al., 2016), which has several variants (Zhang et al., 2015) that are also linear combinations. Yet linear combinations all have one major *flaw*: they give the adversary, described in Section 2.2, full control of the aggregated gradient (and hence of the model parameters), as proven in (Blanchard et al., 2017).

Contrary to linear combinations, the GARs studied here are all proven *Byzantine-resilient* in the sense of Definition 1 introduced in (Blanchard et al., 2017).

All the expected values $\mathbb{E}$ in the following definition and the rest of this paper denote expectations when the source of randomness is the mini-sample $\xi_1 \cdots \xi_{n-f}$ drawn i.i.d by the honest workers.

**Definition 1** (($\alpha, f$)–Byzantine–resilience).

*Let $(\alpha, f) \in [0, \pi/2] \times \{0, \ldots, n\}$ be any (angle, integer), let $n \in \mathbb{N}$ with $n > f$, let $(V_1 \ldots V_{n-f}) \in (\mathbb{R}^d)^{n-f}$ be independent, identically distributed random vectors, with $V_i \sim \mathcal{G}$ and $\mathbb{E}[\mathcal{G}] = G$, and let $(B_1 \ldots B_f) \in (\mathbb{R}^d)^f$ be random vectors, possibly dependent between them and the vectors $(V_1 \ldots V_{n-f})$. Then, an aggregation rule $\mathcal{F}$ is said to be $(\alpha, f)$-Byzantine-resilient if, for any $1 \leq j_1 < \cdots < j_f \leq n$, the vector:*

$$F = \mathcal{F}(V_1, \ldots, \underbrace{B_1}_{j_1}, \ldots, \underbrace{B_f}_{j_f}, \ldots, V_n)$$

*satisfies:*

1. $\langle \mathbb{E}[F], G \rangle \geq (1 - \sin \alpha) \cdot \|G\|^2 > 0$
2. $\forall r \in \{2, 3, 4\}$, $\mathbb{E}\|F\|^r$ *is bounded above by a linear combination of the terms* $\mathbb{E}\|\mathcal{G}\|^{r_1} \cdot \ldots \cdot \mathbb{E}\|\mathcal{G}\|^{r_{n-1}}$, *with* $r_1 + \cdots + r_{n-1} = r$.

Note that the reasoning on expectation holds on the randomness due the $\xi$ samples used to estimate a gradient, the "S" (stochastic) of SGD. Our expectation is not on behaviors of the Byzantine worker, on which we perform a worst case analysis, as should be done in the context of a any safety-critical analysis necessary for Byzantine fault tolerance. Points *1.* and *2.* of this definition are not arbitrary choices: these serve as assumptions for the *global confinement* proof of (Bottou, 1998). Hence, an $(\alpha, f)$–Byzantine–resilient aggregation rule is guaranteed to produce gradients which will make the SGD process *converge*.

### 2.3.1. BRUTE

The *Brute* GAR requires that $n \geq 2f + 1$. Informally, it selects the $n - f$ *most clumped* gradients among the submitted ones, and average them as final output. It is reminiscent of the *Minimal Volume Ellipsoid* estimator, introduced by (Rousseeuw, 1985), and proven to have the optimal *breakdown point* of $50\%$.

Formally, let $(n, f) \in \mathbb{N}^2$ with $n \geq 2f + 1$, let $(V_1 \ldots V_{n-f}) \in (\mathbb{R}^d)^{n-f}$ be independent, identically distributed random vectors, with $V_i \sim \mathcal{G}$ and $\mathbb{E}[\mathcal{G}] = G$, let $(B_1 \ldots B_f) \in (\mathbb{R}^d)^f$ be random vectors, possibly dependent between them and the vectors $(V_1 \ldots V_{n-f})$, let $\mathcal{Q} = \{V_1 \ldots V_n\}$ be the set of submitted gradients, let $\mathcal{R} = \{\mathcal{X} \mid \mathcal{X} \subset \mathcal{Q}, |\mathcal{X}| = n - f\}$ be the set of all the subsets of $\mathcal{Q}$ with a cardinality of $n - f$, and let $\mathcal{S} = \arg\min_{\mathcal{X} \in \mathcal{R}} \left( \max_{(V_i, V_j) \in \mathcal{X}^2} \left( \|V_i - V_j\|_p \right) \right)$.

Then, the aggregated gradient is given by $F = \frac{1}{n-f} \sum_{G \in \mathcal{S}} G$.

Since we use *Brute* as a benchmark when experimenting with small amount of workers, in Section 5.2.1, we also prove its $(\alpha, f)$–Byzantine–resilience. The full proof is available in the supplementary material, Section A.

As a side note, this rule can hardly be used in practical cases, as $|\mathcal{R}| = \frac{n!}{f!(n-f)!}$. For instance, with $n = 57$ workers and $f = 27$, we have $|\mathcal{R}| \approx 1.4 \cdot 10^{16}$. Even with $10^9$ measured subsets $\mathcal{X}$ per second, aggregating these 57 gradients would take more than 5 months.

#### 2.3.2. KRUM

The *Krum* GAR requires that $n \geq 2f + 3$. It is defined in (Blanchard et al., 2017) as follows. Let $(V_1, \ldots, V_n)$ be the $n \geq 2f + 3$ received gradient, among which at most $f$ are Byzantine gradients. For $i \neq j$, let $i \rightarrow j$ denote the fact that $V_j$ belongs to the $n - f - 2$ closest vectors to $V_i$. Finally, let $s(V_i) = \sum_{i \rightarrow j} \|V_i - V_j\|_p^2$ be the *score* of $V_i$. Then, Krum outputs the vector $V_k$ with the lowest score. The $(\alpha, f)$–Byzantine–resilience of Krum is proven by its authors, in the original paper (Blanchard et al., 2017).

#### 2.3.3. THE GEOMETRIC MEDIAN(S) - GEOMED

The third and last gradient aggregation rule we consider is the geometric median (Rousseeuw, 1985). The exact geometric median is known to suffer from computational issues but can be approximated (Cohen et al., 2016). While we know from (Rousseeuw, 1985) that the Median has an optimal breakdown of 0.5, i.e, $n \geq 2f + 1$, it is not known however what would be $\alpha$ such that the Median is $(\alpha, f)$-Byzantine-resilient. Empirically (Blanchard et al., 2017) and theoretically (Chen et al., 2017), variants of the Median were considered as GAR candidates. In particular, the Medoid, which is any minimizer *among* the proposed vectors of the sum of distances, can be used as a GAR and is easier to compute. Since there can be many minimizers, we will simply call GeoMed the Medoid of the proposed vectors with the smallest index.

### 3. Effective attack on $\ell_p$ norm–based GARs

In this section, we describe a simple, yet effective attack passing the Byzantine–resilient GARs, such as the ones presented in Section 2.3. Actually, any $\ell_p$-norm based GAR, where the chosen vector is the result of a distance minimization scheme, is affected.

#### 3.1. Intuition

In high dimensions, the distance function, between two vectors $\|X - Y\|_p$, cannot answer this core question: *do $X$ and $Y$ "disagree" **a bit** on each coordinate, or do they disagree **a lot** on only one?* SGD has proven its ability to accommodate "small errors" from the gradient estimation.

Such "errors" are often *beneficial*, as they may allow the descent process to leave sub–optimal local minima (Bottou, 2012). In Byzantine–free distributed setups, gradient estimations *"disagree" **a bit** on each coordinate*[4].

In a vector space of dimension $d \gg 1$, the "bit of disagreement" on each coordinate translates into a distance $\|X - Y\|_p = \mathcal{O}(\sqrt[p]{d})$. For the omniscient adversary described in Section 2.2, it translates into an opportunity to submit $f$ Byzantine gradients that "disagree" **a lot**, as much as $\mathcal{O}(\sqrt[p]{d})$, on only one coordinate with at least one non–Byzantine gradient. As the $\ell_p$ norm cannot answer the core question mentioned in the above paragraph, such Byzantine gradient could then be *selected* by a $\ell_p$ norm–based GAR.

Each gradient aggregation rule presented in Section 2.3 performs a linear combination of the selected gradient(s). Thus the final aggregated gradient might have one unexpectedly high coordinate. Depending on the learning rate (Figure 4), updating the model with such gradient may push and keep the parameter vector in a sub–space *rarely reached* with the usual, Byzantine–free distributed setup.

The experiments gathered in Section 5.2.1 clearly show this dependency on the learning rate, and indicate that this sub–space only offers sub–optimal to utterly *ineffective* models.

#### 3.2. Attack on the finite norm, $p \geq 1$

The adversary defined in Section 2.2 is omniscient and has arbitrary fast computation and transmission throughput. So for each round, every time the $n - f$ non–Byzantine gradients, are produced, the adversary reads them and chooses the other $f$ gradients the master receives. Based on that capability, for each round, the adversary waits for $n - f$ non–Byzantine gradients to be received. Then it attacks.

Formally, let $\mathcal{Q} = \{V_1 \ldots V_{n-f}\}$ be the set of submitted, non–Byzantine gradients, with $\forall i \in [1 \ldots n - f], V_i \in \mathbb{R}^d$. Let $E = (0 \ldots 0, 1, 0 \ldots 0) \in \mathbb{R}^d$ be any coordinate to attack, and let $\mathcal{B}(\gamma) = \frac{1}{n-f} \left( \sum_{V \in \mathcal{Q}} V \right) + \gamma E$. By a simple linear regression, we estimate the highest value of $\gamma$, noted $\gamma_m$, such that $B = \mathcal{B}(\gamma_m)$ is selected by the aggregation rule. Finally, $B$ is submitted by every Byzantine worker.

For each presented GAR, we reveal a relation between a *rough* estimation of $\gamma_m$ and a few hyper–parameters. We study these approximations of $\gamma_m$ within the *minimal quorum* cases, where the proportion of Byzantine workers is maximized, respectively: $n = 2f + 1$ for Brute and $n = 2f + 3$ for Krum/GeoMed. The full details of the approximations are available in the supplementary material, sections B.2 and B.3. We denote by $\bar{\delta}$ the average *folded* standard deviation on each coordinate of $\mathcal{G}$, and with $p$, $q$ constants: $\gamma_m = \mathcal{O}(\bar{\delta} \sqrt[p]{d})$ for Brute, and $\gamma_m = \mathcal{O}(\bar{\delta} \sqrt[p]{f} \sqrt[p]{d})$ for Krum/GeoMed. The added dependence in $\sqrt[p]{f}$ for Krum/GeoMed comes from the *shape* of the score function, which naturally decreases the score of the Byzantine gradients as they all are identical.

---
[4]This has been observed during the experiments.

As a side–note, an adversary does not necessarily need to know the submitted, non–Byzantine gradients $\mathcal{Q}$ with this attack. Indeed non–Byzantine gradients are assumed to be unbiased, so by the law of large numbers we have: $\lim_{|\mathcal{Q}| \to +\infty} \mathcal{B}(\gamma) = \mathbb{E}[\mathcal{G}] + \gamma E$. It indicates that, for this attack to succeed as well, the adversary may only need to compute an unbiased gradient estimate by itself (without the need to "spy" on the other workers) then add $\gamma E$ to it.

### 3.3. Attack on the infinite norm

In the previous subsection, we have seen that $\gamma_m$ vary as $\sqrt[p]{d}$. Yet, with $d \gg 1$ fixed: $\lim_{p \to +\infty} \sqrt[p]{d} = 1$. Basically, the *curse of dimensionality* exploited in the attack of Section 3.2 no longer exists with $p$ large *enough*, or *infinite*.

One effective attack in the case of an *infinite* norm is simply to change the vector $E = (0 \ldots 0, 1, 0 \ldots 0)$ introduced in the previous subsection for $E = (1 \ldots 1)$. The idea is that modifying non–maximal coordinates of a given vector does not *substantially* affect[5] the distance to the unbiased gradient for the modified vector. From this change on $E$, we proceed as in the previous subsection.

## 4. Bulyan

In addition to being Byzantine–resilient in the sense that it ensures convergence, our algorithm, *Bulyan*[6], also ensures that each coordinate is *agreed on* by a majority of vectors[7] that were selected by a Byzantine–resilient aggregation rule $\mathcal{A}$. This rule $\mathcal{A}$ can for example be *Brute*, *Krum*, a Medoid, the geometric median or any other Byzantine–resilient rule based on an $\ell^p$ norm or the infinite norm.

Let $\mathcal{A}$ be any $(\alpha, f)$-Byzantine–resilient aggregation rule. Bulyan($\mathcal{A}$) requires $n \geq 4f + 3$ received gradients in two steps. The first one is to *recursively* use $\mathcal{A}$ to select $\theta = n - 2f$ gradients, namely:

1. With $\mathcal{A}$, choose, among the proposed vectors, the closest one to $\mathcal{A}$'s output; for Krum or the Medoid, this would be the exact output of $\mathcal{A}$.
2. Remove the chosen gradient from the "received set", and add it to the "selection set", noted $\mathcal{S}$.
3. Loop back to step 1 if $|\mathcal{S}| < \theta$, with $|\cdot|$ the cardinality.

With $n \geq 4f + 3$, we ensure that there is a *quorum* of workers, i.e. $2f + 3$, for each use of $\mathcal{A}$.

Since $\theta = n - 2f \geq 2f + 3$, this selection $\mathcal{S} = (S_1 \ldots S_\theta)$ contains a majority of non–Byzantine gradients. Hence for each $i \in [1..d]$, the median of the $\theta$ coordinates $i$ of the selected gradients is always bounded by coordinates from non–Byzantine submissions. With $\beta = \theta - 2f \geq 3$, the second step is to *generate* the resulting gradient $G = (G[1] \ldots G[d])$, so that for each of its coordinates $G[\cdot]$:

$$\forall i \in [1..d], \; G[i] = \frac{1}{\beta} \sum_{X \in \mathcal{M}[i]} X[i]$$

where: $\mathcal{M}[i] = \underset{\mathcal{R} \subset \mathcal{S}, |\mathcal{R}| = \beta}{\arg\min} \left( \sum_{X \in \mathcal{R}} |X[i] - \text{median}[i]| \right)$

and: $\text{median}[i] = \underset{m = Y[i], Y \in \mathcal{S}}{\arg\min} \left( \sum_{Z \in \mathcal{S}} |Z[i] - m| \right).$

Simply stated: each $i$-th coordinate of $G$ is equal to the average of the $\beta$ closest $i$-th coordinates to the median $i$-th coordinate of the $\theta$ selected gradients.

Let $\mathcal{C}$ be the computational cost of running $\mathcal{A}$ for each epoch at the master to aggregate the gradients.

**Proposition 1.** *(Cost of one Bulyan($\mathcal{A}$) aggregation.)*
*(1) The average computational complexity of Bulyan($\mathcal{A}$) is $\mathcal{O}((n - 2f)\mathcal{C} + dn)$ for each epoch on the master.*
*(2) If $\mathcal{A}$ is GeoMed or Krum, this cost is $\mathcal{O}(n^2 d)$.*

*Proof.* (1) We iterate $\mathcal{A}$ as much as $\theta = n - 2f$ times to get the selected vectors, then we run quick–select to get each median component ($\mathcal{O}(n)$ on each coordinate, i.e. $\mathcal{O}(dn)$ times) and another quick–select to get the $\beta$ closest coordinates (another $\mathcal{O}(dn)$). Note: we use quick–select instead of quick–sort since we do not need ordered values, just the set of the $\beta$ closest values.

For point (2) of the proposition, in fact, if we know more about how $\mathcal{A}$ is performed, we can get rid of the $n - 2f$ multiplications when iterating $\mathcal{A}$: concretely, $\mathcal{A}$ relies on distance computations between proposed vectors, when we iterate it in the same epoch, we do not need to re-compute those distances and would amortize the cost. For instance, for Krum or GeoMed, Bulyan(Krum) and Bulyan(GeoMed) have a cost of $\mathcal{O}(n^2 d + n d)$. In modern machine learning, the models are very large, and $d \gg n$ holds. Therefore Bulyan runs in the same $\mathcal{O}(n^2 d)$ of the base GAR rule it is using[8]. □

**Byzantine Leeway Reduction by Bulyan.**

In the introduction of this paper and in Section 3, we explained that the curse of dimensionality leaves the Byzantine worker, at a coordinate $i$, with a margin of $\Omega(f(d))$ computed as the difference between the Byzantine proposed $i$-th coordinate and the honest proposed vectors' $i$-th coordinates. In what follows, we prove that any vector produced by Bulyan is constrained, in each coordinate, to remain within $\mathcal{O}\left(\frac{\sigma}{\sqrt{d}}\right)$ of honest workers' coordinates. Therefore, narrowing the aforementioned margin to the desired $\mathcal{O}\left(\frac{1}{\sqrt{d}}\right)$.

---

[5] It may not affect the infinite norm at all for small–enough $\gamma$.

[6] Bulyan, Ilyan or more commonly, Julian count of Ceuta, was a Byzantine general, stationed in north Africa, who betrayed the Byzantine empire, in this sense, he was "Byzantine to the Byzantines".

[7] Agreed through the gradients they submitted, of course.

[8] For more comparison, averaging (which is not Byzantine–resilient) already has a cost of $\mathcal{O}(dn)$.

**Proposition 2.** *Denote by $Bu_t$ the vector chosen by Bulyan($\mathcal{A}$) at round t. Then for any dimension $i \in [1, d]$ and any honest worker k proposing gradient $g_k$, we have $\mathbb{E}|Bu_t[i] - g_k[i]| = \mathcal{O}\left(\frac{\sigma}{\sqrt{d}}\right)$.*

*Proof.* Let $\xi = (\xi_1, \cdots, \xi_{n-f})$ denote the random $(n-f)$-tuple of samples used by the honest workers. By assumption, the $\xi_k, k = 1 \cdots n - f$, are assumed to be i.i.d. Let $i \in [1 .. d]$ be any component. We denote by $B$ any vector that is selected by Buylan(A) in the set $\mathcal{M}[i]$ (i.e, $B[i]$ scores among the $\beta$ closest values to $median[i]$). Let $k$ be any honest worker proposing gradient $g_k$, since $B$ was selected by Bulyan, then $B[i]$ is among the closest $\beta$ propositions to $median[i]$. We know that $median[i]$ is the the median coordinate of $\theta \geq 2f + 3$ propositions, and we know that $\beta = \theta - 2f$ therefore, all the set $\mathcal{M}[i]$ is closer to $median[i]$ than at least $2f$ other propositions, in particular, on each side of $median[i]$ (we are in a single dimension) there are at least $f$ workers who are farther from $median[i]$ than is any $B[i]$. Therefore, there are at least two different honest workers, call them $l$ and $r$ whose $i$-th coordinates are respectively on the left and on the right of the $B[i]$, for every $B$ in $\mathcal{M}[i]$, i.e, $g_l[i] \leq B[i] \leq g_r[i]$. There are three cases:

1) $g_k[i] \in ]-\infty, g_l[i]]$, then $|B[i] - g_k[i]| < |g_l[i] - g_k[i]|$
2) $g_k[i] \in ]g_l[i], g_r[i][$, then $|B[i] - g_k[i]| < |g_l[i] - g_r[i]|$
3) $g_k[i] \in [g_r[i], +\infty[$, then $|B[i] - g_k[i]| < |g_r[i] - g_k[i]|$

Denote by $\mathbb{I}_h$ the indicator function of each of the three situations $h = 1, 2, 3$, i.e. $\mathbb{I}_h = 1$ only if we are in case $h$, and $\mathbb{I}_h = 0$ otherwise. Then we have the following bound:

$$|B[i] - g_k[i]| < \mathbb{I}_1 |g_l[i] - g_k[i]| + \mathbb{I}_2 |g_l[i] - g_r[i]| + \mathbb{I}_3 |g_r[i] - g_k[i]|$$

Let $B_1, \cdots, B_\beta$ be the $\beta$ elements of $\mathcal{M}[i]$, the previous inequality holds for every $B_h$, denote by $\mathbb{I}_{r,h}, r = 1 \ldots 3$ the corresponding indicator functions for each $h$, we have:

$$|Bu_t[i] - g_k[i]| \leq \frac{1}{\beta} \sum_{h=1}^{\beta} |B_h[i] - g_k[i]|$$

$$\leq \frac{1}{\beta} \sum_{h=1}^{\beta} (\mathbb{I}_{1,h} |g_l[i] - g_k[i]|$$

$$+ \mathbb{I}_{2,h} |g_l[i] - g_r[i]| + \mathbb{I}_{3,h} |g_r[i] - g_k[i]|)$$

Since $g_l, g_r$ and $g_r$ are all honest workers, which in addition are positioned w.r.t. to other honest workers, they are i.i.d random variables following the randomness of $\xi$ and satisfy a vector–wise variance bound (norm 2) $\mathbb{E}\|g_r - g_l\| = \mathbb{E}\|g_k - g_l\| = \mathbb{E}\|g_k - g_r\| \leq \mathbb{E}\|g_k - G\| + \mathbb{E}\|G - g_r\| = \mathcal{O}(\sigma)$, where $G$ is the unbiased estimator used by the honest workers with a bounded variance such that, component-wise (we divide by $\sqrt{d}$): $\mathbb{E}|Bu_t[i] - g_k[i]| = \mathcal{O}(\frac{\sigma}{\sqrt{d}})$. □

Proposition 2 proves that Bulyan($\mathcal{A}$) reduces the component–wise margin of an attacker, i.e. how much the latter can deviate from honest workers component–wise, while still be influencing the aggregated gradient.

A last natural question to be posed is: will Bulyan($\mathcal{A}$) introduce an additional bias in gradient estimations? The answer, provided by Proposition 1, is *no*. We show that Bulyan($\mathcal{A}$) keeps the gradient estimation in the cone of angle $\alpha$ around the true gradient. In particular, Bulyan($\mathcal{A}$) is also provably convergent.

**Corollary 1.** *Let $\mathcal{A}$ be an $(\alpha, f)$-Byzantine-resilient aggregation rule used by Bulyan. Then Bulyan($\mathcal{A}$) is also $(\alpha, f)$–Byzantine–resilient.*

*Proof.* This is an immediate consequence of the $(\alpha, f)$–Byzantine–resilience of $\mathcal{A}$ (Definition 1) and of the fact that any vector used as an input to the last (averaging) step of Bulyan already comes from the cone of angle $\alpha$, since it was selective by an iteration of $\mathcal{A}$ on a set of vectors of cardinal $\geq 2f + 2$. Let $g$ be the true gradient, a triangle inequality applied between $g$, $Bu$ and the $\beta$ terms coming from the iterations of $\mathcal{A}$, call them $\mathcal{A}_k, k = 1, \cdots \beta$ gives: $\|Bu - g\| \leq \frac{1}{\beta} \|\mathcal{A}_k - g\|$. Given how $\mathcal{A}$'s iterations are performed (without re-sampling $\xi$), the $\mathcal{A}_k$ are themselves i.id and by taking the $\mathbb{E}$ on the inequality, every term in the sum of the right-hand side is bounded by $\|g\| \cdot \sin(\alpha)$ (since it lives in the cone of angle $\alpha$ around $g$. Therefore: $\|\mathbb{E}Bu - g\| \leq \|g\| \cdot \sin(\alpha)$ which means that $\mathbb{E}Bu$ is also a vector in the cone of angle $\alpha$ around $g$. The proof on the statistical moments is obtained with same steps above (except of bounding with $\mathbb{E}\|G\|^r$'s instead of $\sin(\alpha) \cdot \|g\|$ □

Finally, even if the focus of our work was rather on narrowing the leeway of Byzantine workers which we argue is a more powerful requirement than $(\alpha, f)$–Byzantine–resilient alone. It is worth mentioning that as a consequence of our results, convergence is ensured for Bulyan.

**Corollary 2** (Convergence). *With Bulyan($\mathcal{A}$), the sequence of models $x_t$ adopted by the master almost surely converges to a region where $\nabla Q(x) = 0$*

*Proof.* As a consequence of Proposition 1, Bulyan is also $(\alpha, f)$–Byzantine–resilient, by Proposition 2 of (Blanchard et al., 2017) guarantees almost sure convergence. □

# 5. Evaluation

We implemented the three $(\alpha, f)$–Byzantine–resilient gradient aggregation rules presented in Section 2.3, along with the attack introduced in Section 3. We report in this section on the actual impact this attack can have, on the MNIST and CIFAR–10 problems, despite the use of such aggregation rules. Then, we evaluate the impact of Bulyan compared to these gradient aggregation rules. Finally, we exhibit the cost, in terms of *convergence speed*, of using Bulyan in a Byzantine–free setup.

## 5.1. Overview of the studied models

**MNIST.** We use a fully connected, feed–forward network with 784 inputs, 1 hidden layer of size 100, for a total of $d \approx 8 \cdot 10^4$ free parameters. The hidden layers use rectified linear units only. The output layer uses *softmax*.

**CIFAR–10.** We use a convolutional network with the following 7–layers architecture: input $32 \times 32 \times 3$, convolutional (kernel–size: $3 \times 3$, 16 maps, 1 stride), max–pooling of size $3 \times 3$, convolutional (kernel–size: $4 \times 4$, 64 maps, 1 stride), max–pooling of size $4 \times 4$, two fully connected layers composed of 384 and 192 rectified linear units respectively, and *softmax* is used on the output layer. This model totals $10^6$ free parameters. The hidden layers use rectified linear units. The output layer uses *softmax*.

The *maximum cross entropy* loss function is used for both models. L2–regularization of value $10^{-4}$ is used for both models, and both use the Xavier weight initialization algorithm. We use a fading learning rate $\eta\,(epoch) = \eta_0 \frac{r_\eta}{epoch+r_\eta}$. The initial learning rate $\eta_0$, the fading rate $r_\eta$, and the mini–batch size depend on each experiment.

The accuracy is always measured on the *testing set*.

## 5.2. Results

### 5.2.1. ATTACK ON BRUTE, KRUM AND GEOMED

Figures 2 and 3 shows the impact of our attack on the aggregation rules presented in Section 2.3. The *average* rule computes the arithmetic mean of the submitted gradients.

On MNIST, we use $\eta_0 = 1$, $r_\eta = 10000$, a batch size of 83 images (256 for Brute), and for the worker counts:

| | |
|---:|:---|
| Krum/GeoMed | 30 non–Byzantines + 27 Byzantines |
| Brute | 6 non–Byzantines + 5 Byzantines |
| Average | 30 non–Byzantines + 0 Byzantines |

On CIFAR–10, we use $\eta_0 = 0.1$, $r_\eta = 2000$, a batch size of 128 images (256 for Brute), and for the worker counts:

| | |
|---:|:---|
| Krum/GeoMed | 21 non–Byzantines + 18 Byzantines |
| Brute | 6 non–Byzantines + 5 Byzantines |
| Average | 21 non–Byzantines + 0 Byzantines |

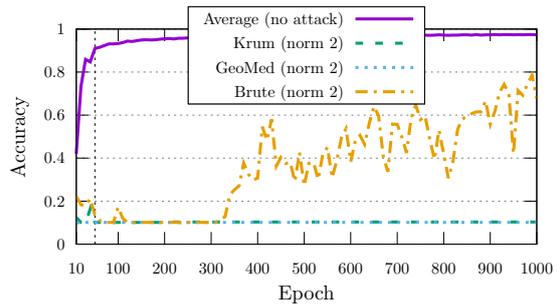

*Figure 2.* MNIST: accuracy on the testing set up to epoch 1000, comparing the presented aggregation rules under our attack. The attack was maintained only up to epoch 50 (dotted line). The *average* is the reference: it is the accuracy the model would have shown if only non–Byzantine gradients had been selected.

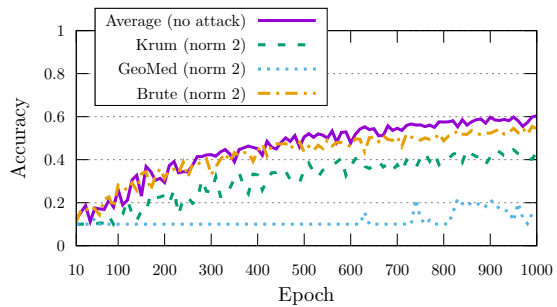

*Figure 3.* CIFAR–10: accuracy on the testing set up to epoch 1000, comparing the presented aggregation rules under our attack. The *average* is the reference: it is the accuracy the model would have shown if only non–Byzantine gradients had been selected.

In Figure 2, the attack is maintained only up to 50 epochs. As shown, and except for Brute, this short attack phase at the beginning of the learning process is sufficient to put the parameter vector in a sub–space of *ineffective* models that SGD did not succeed in leaving for at least 950 epochs. In Figure 3, the attack is never stopped. Only Brute preserved the accuracy. Krum suffered a 33% decrease at epoch 1000, and GeoMed failed to produce a useful model.

Higher learning rates and lower batch sizes naturally extend the effectivity of our attack, by increasing both its *exploratory* capabilities and the variance of the non–Byzantine submissions. The supplementary material, in Section C.1, presents slightly different initial parameters, for which the attack completely prevented any learning.

### 5.2.2. THE EFFECT OF BULYAN

Figures 4 and 5, respectively for MNIST and CIFAR–10, compares Krum, GeoMed and Bulyan (with $\mathcal{A}$ = Krum).

On MNIST, we use $\eta_0 = 1$ ($\eta_0 = 0.2$ for the upper graph), $r_\eta = 10000$, and a mini–batch size of 83 images. On CIFAR–10, we use $\eta_0 = 0.25$, $r_\eta = 2000$, and a mini–batch size of 128 images. For both MNIST and CIFAR–10, we use 30 non–Byzantines + 9 Byzantines workers. Brute cannot be used with that many workers, see Section 2.3.1. In Figure 4, with $\eta_0 = 1$, Krum and GeoMed fail to prevent the attack from *pushing* the model into an *ineffective* state, despite the reduced proportion of Byzantine workers from roughly $1/2$, in Figure 2, to roughly $1/4$. With $\eta_0 = 0.2$, Krum and GeoMed support the attack, at the cost of a *uselessly* slower learning process. Here, Bulyan is not affected by the attack, and achieves the same accuracy *as if* it averages only the non–Byzantine gradients. In Figure 5, we do the same experiment with CIFAR–10. As with MNIST, only Bulyan is not affected by our attack.

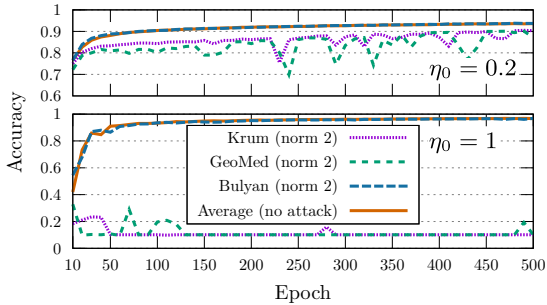

*Figure 4.* MNIST: accuracy on the testing set up to 500 epochs for Krum, GeoMed, Bulyan ($\mathcal{A}$ = Krum) rules. This graph illustrates the impact of the learning rate, as described in Section 3.1.

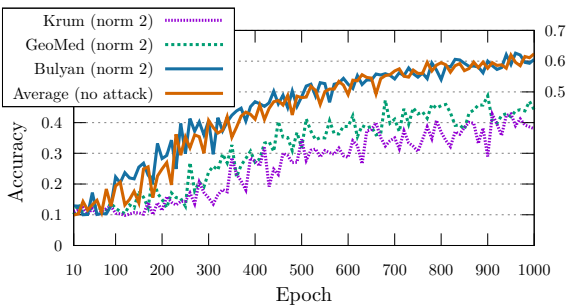

*Figure 5.* CIFAR–10: accuracy on the testing set up to 1000 epochs for Krum, GeoMed, Bulyan ($\mathcal{A}$ = Krum) rules. The arithmetic mean of non–Byzantine gradients serves as reference.

### 5.2.3. THE COST OF BULYAN

For both MNIST and CIFAR–10, we use the same configuration as in the experiments of Section 5.2.2.

In Figure 6, we study the cost of using Bulyan, in terms of *convergence speed*, when there is actually no adversary. We define the *convergence speed*, for a given mini–batch size, as the accuracy the model reaches at a fixed, arbitrary epoch. We use the *average*, i.e. the arithmetic mean of the submitted gradients, as the reference aggregation rule.

Without Byzantine workers, the loss in convergence speed induced by Bulyan is minimized with a *reasonable* batch size: 24 images/batch for MNIST, and 36 for CIFAR–10.

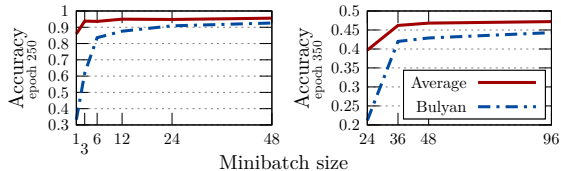

*Figure 6.* MNIST (left), CIFAR–10 (right): Accuracy on the testing set at epoch 250 for Average and Bulyan. There are $n = 39$ workers and no adversary, but $f$ is declared to 9. This shows the trade–off between the Byzantine robustness of Bulyan and the loss in convergence speed it introduces.

## 6. Concluding remarks

At first glance, the Byzantine–failure model might seem specific to distributed computing, and too pessimistic in a general ML context, for it assumes that some workers actively try to fool SGD towards the worst possible direction. In fact, we argue that this failure model is also relevant for poisoning attacks on single machines, as well as learning with unreliable data in a centralized, single–worker setting.

For instance, even when there is no *group of workers* whose estimated gradients are aggregated, an SGD update is still an attempt to *aggregate* knowledge from data and update the model subsequently. Consider a centralized, single learner, drawing data from a distribution containing a fraction of corrupt samples. The 0.5 breakdown limit for unbiased estimators, initially formulated (Rousseeuw, 1985) without computability considerations, establishes a formal limit to poisoning attacks: they are impossible to counter if at least half of the data is not i.i.d from the (desired) training distribution, even with a convex cost function in small dimensions. In very high dimensions, and with highly–non convex cost functions, our work shows that the inaugural distributed–computing defenses (Blanchard et al., 2017; Chen et al., 2017; Su, 2017) against poisoning attacks, tough provably converging, remain frail in the face of *curse of dimensionality* attacks. They might indeed converge, as promised, but to the worst possible region. To defend against that, we introduce Bulyan, which we theoretically prove to significantly reduce the adversarial leeway that causes this drift to sub–optimal models.

We empirically show that Bulyan, indeed avoids convergence to ineffectual models, and instead, ends up learning models that are comparable to a reasonable benchmark: a non–attacked averaging scheme. However, the question of finding "the best direction" possible for non–convex cost functions remains one of the most challenging ones in optimization for machine learning (Choromanska et al., 2015b), especially when we stick to such a cheap, first–order method as SGD, and avoid expensive, Hessian–like, computations as pointed out by (Reddi et al., 2017).

It is important to recall that our approach tackles what happens *while* training, with the goal of avoiding bad models due to poisoning attacks. We did not address the problem of evasion attacks, i.e. attacking an already trained model with adversarial inputs.

From a distributed computing point of view, there seems to be two complementary views on robustness to be made: (1) a coarse-grained robustness (quality of the gradient aggregation scheme), and (2) a fine-grained robustness (quality of the model w.r.t single parameters). For the first view, the unit of failure would be a single machine, hosting a copy of the model (or significant parts of it) and attacked in its attempt to estimate gradients. For the second view, the models used by ML, neural networks for example, can themselves be viewed as distributed computing objects (Piuri, 2001), where the units of failure are individual neurons and weight values. The question was explored (Kerlirzin, 1994) in the 1990s with unexpected connections with todays' popular tools such as the dropout algorithm, which was initially derived with a (distributed computing) fault-tolerance purpose (Kerlirzin & Vallet, 1993), then independently re-discovered as a robust regularization scheme to reduce over-fitting 20 years later (Srivastava et al., 2014). Recently, this gap-bridging between robustness from the distributed computing point of view, and robustness, from a learning performance perspective is being revived, with questions such as: how much error will be *forward propagated* if units of the model are erroneous?

Recently, NVIDIA investigated (a closely related statement of) the question through experimental lenses (Li et al., 2017). In the mean time, theoretical evaluations of this propagated error were derived with a tight bound (El Mhamdi & Guerraoui, 2016; 2017; El Mhamdi et al., 2017) which relies on computing the Lipschitz coefficient of the neural network and proving its exponential dependency with the depth. A similar (exponential in the depth) pattern has been derived while investigating robustness (to evasion attacks (Cisse et al., 2017)). We argue that poisoning attacks should be studied both from a fine-grained, and coarse-grained perspective. For instance, the curse of dimensionality attack (Section 3) is an example of a coarse-grained problem (the whole system being wrong about the model) arising from a fine-grained (single-parameter) attack. An interesting question can therefore be posed: given bounds on how much an error in an individual weight value can influence the output of a model, can we compute the impact of a poisoning attack of one single component of the gradient? Subsequently, could this link be leveraged to discover the theoretical limits of *any* defense against poisoning?

## References

Abadi, Martín, Barham, Paul, Chen, Jianmin, Chen, Zhifeng, Davis, Andy, Dean, Jeffrey, Devin, Matthieu, Ghemawat, Sanjay, Irving, Geoffrey, Isard, Michael, et al. Tensorflow: A system for large-scale machine learning. In *Proceedings of the 12th USENIX Symposium on Operating Systems Design and Implementation (OSDI). Savannah, Georgia, USA*, 2016.

Biggio, Battista and Laskov, Pavel. Poisoning attacks against support vector machines. In *In International Conference on Machine Learning (ICML.* Citeseer, 2012.

Biggio, Battista and Roli, Fabio. Wild patterns: Ten years after the rise of adversarial machine learning. *arXiv preprint arXiv:1712.03141*, 2017.

Blanchard, Peva, El Mhamdi, El Mahdi, Guerraoui, Rachid, and Stainer, Julien. Machine learning with adversaries: Byzantine tolerant gradient descent. In *Advances in Neural Information Processing Systems 30*, pp. 118–128. Curran Associates, Inc., 2017.

Bojarski, Mariusz, Del Testa, Davide, Dworakowski, Daniel, Firner, Bernhard, Flepp, Beat, Goyal, Prasoon, Jackel, Lawrence D, Monfort, Mathew, Muller, Urs, Zhang, Jiakai, et al. End to end learning for self-driving cars. *arXiv preprint arXiv:1604.07316*, 2016.

Bottou, Léon. Online learning and stochastic approximations. *Online learning in neural networks*, 17(9):142, 1998.

Bottou, Léon. *Stochastic Gradient Descent Tricks*, pp. 421–436. Springer Berlin Heidelberg, Berlin, Heidelberg, 2012. ISBN 978-3-642-35289-8. doi: 10.1007/978-3-642-35289-8_25. URL https://doi.org/10.1007/978-3-642-35289-8_25.

Chen, Yudong, Su, Lili, and Xu, Jiaming. Distributed statistical machine learning in adversarial settings: Byzantine gradient descent. *arXiv preprint arXiv:1705.05491*, 2017.

Choromanska, Anna, Henaff, Mikael, Mathieu, Michael, Ben Arous, Gérard, and LeCun, Yann. The loss surfaces of multilayer networks. In *Artificial Intelligence and Statistics*, pp. 192–204, 2015a.

Choromanska, Anna, LeCun, Yann, and Ben Arous, Gérard. Open problem: The landscape of the loss surfaces of multilayer networks. In *Conference on Learning Theory*, pp. 1756–1760, 2015b.

Cisse, Moustapha, Bojanowski, Piotr, Grave, Edouard, Dauphin, Yann, and Usunier, Nicolas. Parseval networks: Improving robustness to adversarial examples. In *International Conference on Machine Learning*, pp. 854–863, 2017.

Cohen, Michael B, Lee, Yin Tat, Miller, Gary, Pachocki, Jakub, and Sidford, Aaron. Geometric median in nearly linear time. In *Proceedings of the forty-eighth annual ACM symposium on Theory of Computing*, pp. 9–21. ACM, 2016.

Dean, Jeffrey, Corrado, Greg, Monga, Rajat, Chen, Kai, Devin, Matthieu, Mao, Mark, Senior, Andrew, Tucker, Paul, Yang, Ke, Le, Quoc V, et al. Large scale distributed deep networks. In *Advances in neural information processing systems*, pp. 1223–1231, 2012.

El Mhamdi, El Mahdi and Guerraoui, Rachid. When neurons fail. In *2017 IEEE International Parallel and Distributed Processing Symposium (IPDPS)*, pp. 1028–1037, May 2017. doi: 10.1109/IPDPS.2017.66.

El Mhamdi, El Mahdi, Guerraoui, Rachid, and Rouault, Sébastien. On the robustness of a neural network. In *2017 IEEE 36th Symposium on Reliable Distributed Systems (SRDS)*, pp. 84–93, Sept 2017. doi: 10.1109/SRDS.2017.21.

El Mhamdi, El Mahdi and Guerraoui, Rachid. When neurons fail - technical report. pp. 19, 2016. Biological Distributed Algorithms Workshop, Chicago.

Gilmer, Justin, Metz, Luke, Faghri, Fartash, Schoenholz, Samuel S, Raghu, Maithra, Wattenberg, Martin, and Goodfellow, Ian. Adversarial spheres. *arXiv preprint arXiv:1801.02774*, 2018.

Goodfellow, Ian J, Shlens, Jonathon, and Szegedy, Christian. Explaining and harnessing adversarial examples. *arXiv preprint arXiv:1412.6572*, 2014.

Holzinger, Andreas. Interactive machine learning for health informatics: when do we need the human-in-the-loop? *Brain Informatics*, 3(2):119–131, 2016.

Kerlirzin, P and Vallet, F. Robustness in multilayer perceptrons. *Neural computation*, 5(3):473–482, 1993.

Kerlirzin, Philippe. *Etude de la robustesse des réseaux multicouches*. PhD thesis, Paris 11, 1994.

Koh, Pang Wei and Liang, Percy. Understanding blackbox predictions via influence functions. In *International Conference on Machine Learning*, pp. 1689–1698, 2017.

Kumar, Atul, Mehta, Sameep, and Vijaykeerthy, Deepak. An introduction to adversarial machine learning. In *International Conference on Big Data Analytics*, pp. 293–299. Springer, 2017.

Lamport, Leslie, Shostak, Robert, and Pease, Marshall. The byzantine generals problem. *ACM Transactions on Programming Languages and Systems (TOPLAS)*, 4(3): 382–401, 1982.

LeCun, Yann, Bengio, Yoshua, and Hinton, Geoffrey. Deep learning. *Nature*, 521(7553):436–444, 2015.

Li, Guanpeng, Hari, Siva Kumar Sastry, Sullivan, Michael, Tsai, Timothy, Pattabiraman, Karthik, Emer, Joel, and Keckler, Stephen W. Understanding error propagation in deep learning neural network (dnn) accelerators and applications. 2017.

Li, Mu, Andersen, David G, Park, Jun Woo, Smola, Alexander J, Ahmed, Amr, Josifovski, Vanja, Long, James, Shekita, Eugene J, and Su, Bor-Yiing. Scaling distributed machine learning with the parameter server. In *OSDI*, volume 1, pp. 3, 2014a.

Li, Mu, Andersen, David G, Smola, Alexander J, and Yu, Kai. Communication efficient distributed machine learning with the parameter server. In *Advances in Neural Information Processing Systems*, pp. 19–27, 2014b.

Piuri, Vincenzo. Analysis of fault tolerance in artificial neural networks. *Journal of Parallel and Distributed Computing*, 61(1):18–48, 2001.

Reddi, Sashank J, Zaheer, Manzil, Sra, Suvrit, Poczos, Barnabas, Bach, Francis, Salakhutdinov, Ruslan, and Smola, Alexander J. A generic approach for escaping saddle points. *arXiv preprint arXiv:1709.01434*, 2017.

Rousseeuw, Peter J. Multivariate estimation with high breakdown point. *Mathematical statistics and applications*, 8:283–297, 1985.

Srivastava, Nitish, Hinton, Geoffrey, Krizhevsky, Alex, Sutskever, Ilya, and Salakhutdinov, Ruslan. Dropout: A simple way to prevent neural networks from overfitting. *The Journal of Machine Learning Research*, 15(1): 1929–1958, 2014.

Su, Lili. *Defending distributed systems against adversarial attacks: consensus, consensus-based learning, and statistical learning*. PhD thesis, University of Illinois at Urbana-Champaign, 2017.

Zhang, Sixin, Choromanska, Anna E, and LeCun, Yann. Deep learning with elastic averaging sgd. In *Advances in Neural Information Processing Systems*, pp. 685–693, 2015.

The Hidden Vulnerability of Distributed Learning in Byzantium# A. Brute's $(\alpha, f)$–Byzantine–resilience proof

## A.1. Background

**Definition 2** (($\alpha, f$)–Byzantine–resilience).
Let $(n, f) \in (\mathbb{N}^*)^2$ with $n > f$.
Let $(\alpha, f) \in [0, \pi/2] \times [0..n]$ be any angle and any integer.
Let $(V_1 \ldots V_{n-f}) \in (\mathbb{R}^d)^{n-f}$ be independent, identically distributed random vectors, with $V_i \sim \mathcal{G}$ and $\mathbb{E}[\mathcal{G}] = G$.
Let $(B_1 \ldots B_f) \in (\mathbb{R}^d)^f$ be random vectors, possibly dependent between them and the vectors $(V_1 \ldots V_{n-f})$.
Then, an aggregation rule $\mathcal{F}$ is said to be $(\alpha, f)$-Byzantine-resilient if, for any $1 \leq j_1 < \cdots < j_f \leq n$, the vector:

$$F = \mathcal{F}\left(V_1, \ldots, \underbrace{B_1}_{j_1}, \ldots, \underbrace{B_f}_{j_f}, \ldots, V_n\right)$$

satisfies:

1. $\langle \mathbb{E}[F], G \rangle \geq (1 - \sin \alpha) \cdot \|G\|^2 > 0$
2. $\forall r \in \{2, 3, 4\}$, $\mathbb{E}\|F\|^r$ is bounded above by a linear combination of the terms $\mathbb{E}\|\mathcal{G}\|^{r_1} \cdot \ldots \cdot \mathbb{E}\|\mathcal{G}\|^{r_{n-1}}$, with $r_1 + \cdots + r_{n-1} = r$.

## A.2. Definition

Let $(n, f) \in (\mathbb{N}^*)^2$ with $n \geq 2f + 1$.
Let $(V_1 \ldots V_{n-f}) \in (\mathbb{R}^d)^{n-f}$ be independent, identically distributed random vectors, with $V_i \sim \mathcal{G}$ and $\mathbb{E}[\mathcal{G}] = G$.
Let $(B_1 \ldots B_f) \in (\mathbb{R}^d)^f$ be random vectors, possibly dependent between them and the vectors $(V_1 \ldots V_{n-f})$.
Let $\|\cdot\|_p$ be the $\ell_p$–norm, with $p \in \mathbb{N}^* \cup \{+\infty\}$.
Let $\mathcal{Q} = \{V_1 \ldots V_n\}$ be the set of submitted gradients.
Let $\mathcal{R} = \{\mathcal{X} \mid \mathcal{X} \subset \mathcal{Q}, |\mathcal{X}| = n - f\}$ be the set of all the subsets of $\mathcal{Q}$ with a cardinality of $n - f$.
Let $\mathcal{S} = \arg\min_{\mathcal{X} \in \mathcal{R}}\left(\max_{(V_i, V_j) \in \mathcal{X}^2}\left(\|V_i - V_j\|_p\right)\right)$.
Then, the aggregated gradient $F = \frac{1}{n-f}\sum_{V \in \mathcal{S}} V$.

## A.3. Proof

Let $\forall (i, j) \in [1..n - f]^2, i \neq j$ be $\bar{\sigma} \triangleq \mathbb{E}\|V_i - V_j\|_p$.
Under the assumption that $2 f \bar{\sigma} < (n - f) \|G\|_p$, we will prove that this rule is $(\alpha, f)$–Byzantine–resilient.

Trivial case: $\forall i \in [1..f], B_i \notin \mathcal{S}$.
As the aggregated gradient $F$ is the arithmetic mean of unbiased vectors $V_j$, we have $\mathbb{E}[F] = G$, and points *1.* and *2.* of definition 2 are trivially satisfied.

Otherwise, without loss of generality, let $b \in [1..f]$ and $\mathcal{S} = \{V_1 \ldots V_{n-f-b}, B_1 \ldots B_b\}$, $\bar{\mathcal{R}} = \mathcal{R} \setminus \mathcal{S}$. It holds:

$$\forall \bar{\mathcal{S}} \in \bar{\mathcal{R}}, \exists X_i \in \bar{\mathcal{S}} \setminus \mathcal{S}, \exists X_j \in \bar{\mathcal{S}} \setminus \{X_i\},$$

$$\forall X_k \in \mathcal{S}, \forall X_l \in \mathcal{S} \setminus \{X_k\},$$
$$\|X_k - X_l\|_p < \|X_i - X_j\|_p$$

We can also notice that: $\exists \mathcal{V} \in \bar{\mathcal{R}}, \forall i \in [1..f], B_i \notin \mathcal{V}$.
Then, by combining this observation with the previous one:

$$\forall a \in [1..b], B_a \in \mathcal{S}$$
$$\Rightarrow \exists (x, y) \in [1..n - f]^2, i \neq j,$$
$$\forall k \in [1..n - f - b], \|B_a - V_k\|_p < \|V_x - V_y\|_p$$

This last observation will be reused in the following.

We can compute the aggregated gradient:

$$F = \frac{1}{n - f}\left(\sum_{i=1}^{n-f-b} V_i + \sum_{i=1}^{b} B_i\right)$$

and compare it with the average of the non–Byzantine ones:

$$\widehat{G} = \frac{1}{n - f}\sum_{i=1}^{n-f} V_i$$

$$F - \widehat{G} = \frac{1}{n - f}\left(\sum_{i=1}^{b} B_i - \sum_{i=n-f-b+1}^{n-f} V_i\right)$$

$$= \frac{1}{n - f}\sum_{i=1}^{b} B_i - V_{i+n-f-b}$$

$$\left\|F - \widehat{G}\right\|_p \leq \frac{1}{n - f}\sum_{i=1}^{b} \|B_i - V_{i+n-f-b}\|_p$$

$$\leq \frac{1}{n - f}\sum_{i=1}^{b}\left(\|B_i - V_k\|_p\right.$$

$$\left. + \|V_k - V_{i+n-f-b}\|_p\right)$$

$$\leq \frac{1}{n - f}\sum_{i=1}^{b}\left(\|V_x - V_y\|_p\right.$$

$$\left. + \|V_k - V_{i+n-f-b}\|_p\right)$$

We can then compute the expected value of this distance, and with $\mathbb{E}\left[\widehat{G}\right] \triangleq G$ and the Jensen's inequality:

$$\|\mathbb{E}[F] - G\|_p \leq \mathbb{E}\left\|F - \widehat{G}\right\|_p$$

$$\leq \frac{1}{n - f}\sum_{i=1}^{b} \bar{\sigma} + \bar{\sigma}$$

$$\leq \frac{2 b \bar{\sigma}}{n - f} \leq \frac{2 f \bar{\sigma}}{n - f}$$

So, under the assumption that $2f\bar{\sigma} < (n-f)\|G\|_p$, we verify that $\|\mathbb{E}[F] - G\|_p < \|G\|_p$, and so: $\langle \mathbb{E}[F], G \rangle > 0$.

Point 2. can also be verified formally, $\forall r \in \{2, 3, 4\}$:

$$\mathbb{E}\|F\|_p^r \leq \frac{n-f-b}{n-f}\mathbb{E}\|\mathcal{G}\|_p^r + \frac{1}{n-f}\sum_{i=1}^b \mathbb{E}\|B_i\|_p^r$$

Then, by using the binomial theorem twice:

$$\|B_i\|_p^r \leq \sum_{r_1+r_2=r} \binom{r}{r_1} \|B_i - V_k\|_p^{r_1} \|V_k\|_p^{r_2}$$
$$\text{with } k \in [1 \mathinner{.\,.} n - f - d]$$
$$\|B_i - V_k\|_p^{r_1} \leq \|V_x - V_y\|_p^{r_1}$$
$$\leq \sum_{r_3+r_4=r_1} \binom{r_1}{r_3} \|V_x\|_p^{r_3} \|V_y\|_p^{r_4}$$

Finally, as $(V_1 \ldots V_{n-f})$ are *independent, identically distributed* random variables following the same distribution $\mathcal{G}$, we have that $\forall (i,j) \in [1 \mathinner{.\,.} n-f]^2, i \neq j$, $\mathbb{E}\left[\|V_i\|_p^{r_1} \|V_j\|_p^{r_2}\right] = \mathbb{E}\|\mathcal{G}\|_p^{r_1} \cdot \mathbb{E}\|\mathcal{G}\|_p^{r_2}$, and so $\mathbb{E}\|B_i\|_p^r$ is bounded as described in point 2. of definition 2.

# B. Approximation of $\alpha_m$, with $p \in \mathbb{N}^*$

## B.1. Prior conventions and assumptions

Let remind that $\forall i \in [1 \mathinner{.\,.} n-f], V_i = \left(v_1^{(i)} \ldots v_d^{(i)}\right) \sim \mathcal{G}$.

We model each coordinate as a *normal distribution*:

$$\forall j \in [1 \mathinner{.\,.} d], \exists (\mu_j, \sigma_j) \in \mathbb{R}^2,$$
$$\forall i \in [1 \mathinner{.\,.} n-f], v_j^{(i)} \sim \mathcal{N}(\mu_j, \sigma_j^2)$$

We assume $d \gg 1$, and we will write $\bar{\delta}$ for:

$$\forall (i,j) \in [1 \mathinner{.\,.} n-f]^2, i \neq j, \bar{\delta} = \frac{1}{d}\sum_{k=1}^d \mathbb{E}\left|v_k^{(i)} - v_k^{(j)}\right|$$
$$= \frac{2}{d\sqrt{\pi}}\sum_{k=1}^d \sigma_k$$

and note that: $\frac{1}{d}\sum_{k=1}^d \mathbb{E}\left|v_k^{(i)} - \mu_k\right| = \frac{\sqrt{2}}{d\sqrt{\pi}}\sum_{k=1}^d \sigma_k$
$$= \frac{\bar{\delta}}{\sqrt{2}}$$

Then, $\forall (i,j) \in [1 \mathinner{.\,.} n-f]^2, i \neq j$, we can approximate:

$$\|V_i - V_j\|_p = \left(\sum_{k=1}^d \left|v_k^{(i)} - v_k^{(j)}\right|^p\right)^{\frac{1}{p}}$$
$$\approx \left(d\,\bar{\delta}^p\right)^{\frac{1}{p}}$$

Let $E = (0 \ldots 0, 1, 0 \ldots 0) \in \mathbb{R}^d$ the attacked coordinate. Then, with $\alpha_m > 0$, $B = \bar{V} + \alpha_m E$, we can approximate:

$$\|B - V_i\|_p = \left(\left(\sum_{k=1}^d \left|v_k^{(i)} - \bar{v}_k\right|^p\right) - \left|v_e^{(i)} - \bar{v}_e\right|^p + \left|v_e^{(i)} - \bar{v}_e + \alpha_m\right|^p\right)^{\frac{1}{p}}$$
$$\approx \left(\alpha_m^p + \sum_{k=1}^d \left|v_k^{(i)} - \mu_k\right|^p\right)^{\frac{1}{p}}$$
$$\approx \left(\alpha_m^p + d\left(\frac{\bar{\delta}}{\sqrt{2}}\right)^p\right)^{\frac{1}{p}}$$

## B.2. Attack against Brute

We only study the *worst case* scenario, where $n = 2f + 1$, maximizing the proportion of Byzantine workers.

Assuming $B$ is selected by Brute:

$$B \in \mathcal{S}$$
$$\Rightarrow \exists (x,y) \in [1 \mathinner{.\,.} n-f]^2, i \neq j,$$
$$\forall k \in [1 \mathinner{.\,.} n-f-b], \|B - V_k\|_p < \|V_x - V_y\|_p$$
$$\rightsquigarrow \left(\alpha_m^p + d\left(\frac{\bar{\delta}}{\sqrt{2}}\right)^p\right)^{\frac{1}{p}} < \left(d\,\bar{\delta}^p\right)^{\frac{1}{p}}$$
$$\rightsquigarrow \alpha_m < \left(\left(1 - \frac{1}{\sqrt{2}^p}\right)d\right)^{\frac{1}{p}}\bar{\delta}$$

This is a *necessary*, approximated condition. It is only to give broad insights on the relation between some hyper–parameters and $\alpha_m$: with $p, q$ constants, $\alpha_m = \mathcal{O}\left(\bar{\delta}\sqrt[p]{d}\right)$.

## B.3. Attack against Krum/GeoMed

We only study the *worst case* scenario, where $n = 2f + 3$, maximizing the proportion of Byzantine workers.
Let $q \in \{1, 2\}$, $q = 1$ for GeoMed and $q = 2$ for Krum.

First, we approximate the Byzantine submission's score:

$$s(B) \approx 2\|B - V_i\|_p^q$$
$$\approx 2\left(\alpha_m^p + d\left(\frac{\bar{\delta}}{\sqrt{2}}\right)^p\right)^{\frac{q}{p}}$$

$\forall i \in [1 \mathinner{.\,.} n-f]$, let $b \in [0 \mathinner{.\,.} f]$ be how many $B$ belongs to the $n - f - 2$ closest vectors to $V_i$. Then the score of $V_i$ is:

$$s(V_i) \approx b\|B - V_i\|_p^q + (f + 1 - b)\|V_j - V_i\|_p^q$$

$$\approx b \left( \alpha_m{}^p + d \left( \frac{\bar{\delta}}{\sqrt{2}} \right)^p \right)^{\frac{q}{p}} + (f + 1 - b) \left( d\,\bar{\delta}^p \right)^{\frac{q}{p}}$$

Finally, $B$ is selected $\Rightarrow \forall i \in [1\mathbin{..} n - f],\ s(B) \lessapprox s(V_i)$

$$\underset{\forall i}{\overset{\uparrow}{\Rightarrow}}\ (2 - b)\left( \alpha_m{}^p + d \left( \frac{\bar{\delta}}{\sqrt{2}} \right)^p \right)^{\frac{q}{p}} \lessapprox (f + 1 - b)\left( d\,\bar{\delta}^p \right)^{\frac{q}{p}}$$

$$\underset{\exists i}{\overset{\uparrow}{\Rightarrow}}\ \alpha_m \lessapprox \left( \left( \frac{f + 1 - b}{2 - b} \right)^{\frac{p}{q}} - \frac{1}{\sqrt{2}^p} \right)^{\frac{1}{p}} d^{\frac{1}{p}}\,\bar{\delta}$$

This last implication is always true: there *must* be at least one non–Byzantine vector $V_j$ for with $b \in \{0, 1\}$; else $\alpha_m$ could increase unbounded, which would be absurd.

In conclusion, with $p, q$ constants: $\alpha_m = \mathcal{O}\!\left( \bar{\delta} \sqrt[q]{f}\ \sqrt[p]{d} \right)$.

## C. Supplementary experiments

### C.1. Attack on Brute, Krum and GeoMed

On MNIST, here we use $\eta_0 = 1$, $r_\eta = 10000$, a batch size of 83 images (256 for Brute), and for the workers:

| | |
|---:|:---|
| Krum/GeoMed | 30 non–Byzantines + 27 Byzantines |
| Brute | 6 non–Byzantines + 5 Byzantines |
| Average | 30 non–Byzantines + 0 Byzantines |

On CIFAR–10, we use $\eta_0 = 0.5$, $r_\eta = 2000$, a batch size of 128 images (256 for Brute), and for the worker counts:

| | |
|---:|:---|
| Krum/GeoMed | 21 non–Byzantines + 18 Byzantines |
| Brute | 6 non–Byzantines + 5 Byzantines |
| Average | 21 non–Byzantines + 0 Byzantines |

In Figure 7, the attack is maintained only up to 50 epochs. The attack variant for $\ell_\infty$ norm–based gradient aggregation rules exhibited a very strong impact. None of the presented gradient aggregation rules prevented the stochastic gradient descent from being *pushed* and remaining in a sub–space of *ineffective* models, and for at least 1000 epochs.

In Figure 8, the attack is never stopped. Again, none of the presented gradient aggregation rules prevented the stochastic gradient descent from being *pushed* and remaining in a sub–space of *ineffective* models, for at least 1000 epochs.

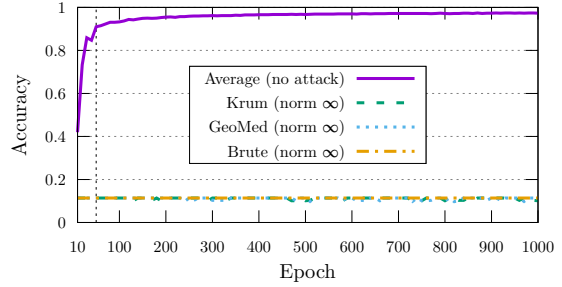

*Figure 7.* MNIST: accuracy on the testing set up to epoch 1000, comparing the presented aggregation rules under our attack. The attack was maintained only up to epoch 50 (dotted line). The *average* is the reference: it is the accuracy the model would have shown if only non–Byzantine gradients had been selected.

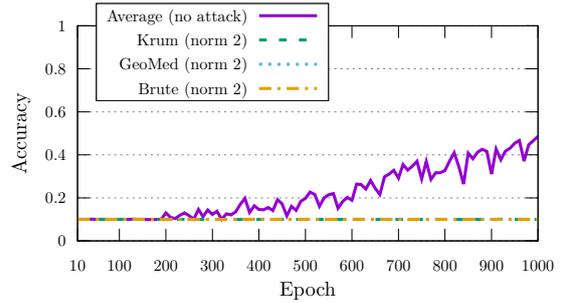

*Figure 8.* CIFAR–10: accuracy on the testing set up to epoch 1000, comparing the presented aggregation rules under our attack. The *average* is the reference: it is the accuracy the model would have shown if only non–Byzantine gradients had been selected.